\documentclass[10pt,twocolumn,letterpaper]{article}

\usepackage[pagenumbers]{cvpr} % To force page numbers, e.g. for an arXiv version

\makeatletter
\@namedef{ver@everyshi.sty}{}
\makeatother

\usepackage{graphicx}
\usepackage{amsmath}
\usepackage{amssymb}
\usepackage{booktabs}

\usepackage{array}
\usepackage{siunitx}
\usepackage{multirow}
\usepackage{todonotes}
\usepackage[pagebackref,breaklinks,colorlinks]{hyperref}

\usepackage[capitalize]{cleveref}
\crefname{section}{Sec.}{Secs.}
\Crefname{section}{Section}{Sections}
\Crefname{table}{Table}{Tables}
\crefname{table}{Tab.}{Tabs.}

\def\cvprPaperID{*****} % *** Enter the CVPR Paper ID here
\def\confName{CVPR}
\def\confYear{2022}

\begin{document}

\title{Semantic Pose Verification for Outdoor Visual Localization with \\Self-supervised Contrastive Learning}

%%%%%%%%% Author Style V2 %%%%%%%%%
 \author{Semih Orhan$^1$, \quad  Jose J. Guerrero$^2$, \quad Yalin Bastanlar$^1$\\
 $^1${Department of Computer Engineering, Izmir Institute of Technology}\\
 \{semihorhan,yalinbastanlar\}@iyte.edu.tr\\
 $^2${Instituto de Investigación en Ingeniería de Aragón (I3A), Universidad de Zaragoza}\\
 {jguerrer@unizar.es}
}
 \maketitle
%%%%%%%%% --------------- %%%%%%%%%

%%%%%%%%% ABSTRACT
\begin{abstract}
   Any city-scale visual localization system has to overcome long-term appearance changes, such as varying illumination conditions or seasonal changes between query and database images. Since semantic content is more robust to such changes, we exploit semantic information to improve visual localization. In our scenario, the database consists of gnomonic views generated from panoramic images (e.g. Google Street View) and query images are collected with a standard field-of-view camera at a different time. To improve localization, we check the semantic similarity between query and database images, which is not trivial since the position and viewpoint of the cameras do not exactly match. To learn similarity, we propose training a CNN in a self-supervised fashion with contrastive learning on a dataset of semantically segmented images. With experiments we showed that this semantic similarity estimation approach works better than measuring the similarity at pixel-level. Finally, we used the semantic similarity scores to verify the retrievals obtained by a state-of-the-art visual localization method and observed that contrastive learning-based pose verification increases top-1 recall value to 0.90 which corresponds to a 2\% improvement. 
\end{abstract}

%%%%%%%%% BODY TEXT
\section{Introduction}
\label{sec:intro}

Visual localization can be defined as estimating the position of a visual query material within a known environment. It has received increasing attention \cite{piasco2018survey,arandjelovic2016netvlad,ge2020self,chen2017deep,jegou2011aggregating, taira2019right} in the last decade especially due to the limitation of GPS-based localization in urban environment (e.g. signal failure in cluttered environment) and motivated by many computer vision application areas such as autonomous vehicle localization \cite{toft2020long}, unmanned aerial vehicle localization \cite{couturier2021review}, virtual and augmented reality \cite{lynen2020large}.

Visual localization technique that we employ is based on image retrieval, where query images are searched within a geo-tagged database. Location of the retrieved database image serves as the estimated position of the query image. Both query and database images are represented with compact and distinguishable fixed size set of features \cite{arandjelovic2016netvlad, ge2020self, radenovic2018fine,radenovic2018revisiting,tolias2016particular,azizpour2015generic}. In recent years, features extracted with convolutional neural networks (CNNs) \cite{arandjelovic2016netvlad,ge2020self,radenovic2018fine} outperformed hand-crafted features \cite{arandjelovic2012three,philbin2007object,jegou2011aggregating}.

In our work, query images are collected with a standard field-of-view camera \cite{zamir2014image}, whereas database consists of perspective images (gnomonic projection) generated from a panoramic image dataset (downloaded from Google Street View). The reason for using the panoramic image database is that it presents a wide field-of-view (FOV) which helps to correctly localize the query images where standard FOV cameras fail due to non-overlapping fields of view. 

\begin{figure}[t]
    \centering
    \includegraphics[scale=0.17]{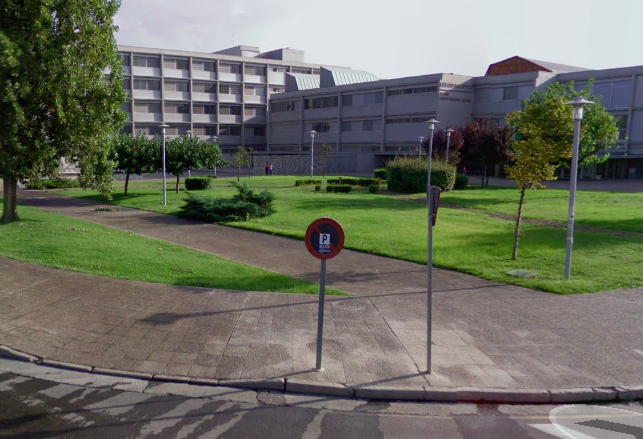}
    \includegraphics[scale=0.17]{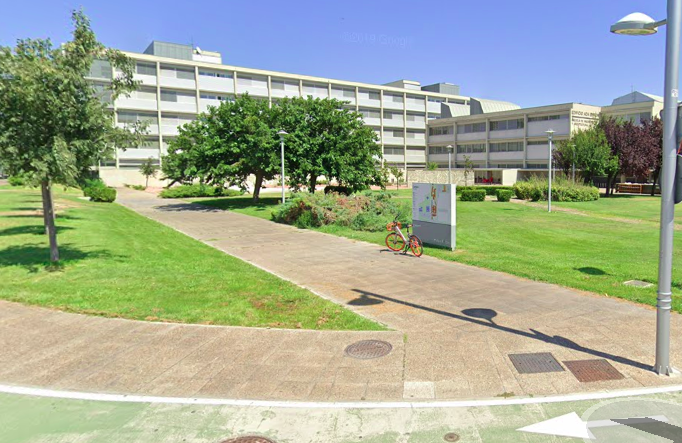}
    \includegraphics[scale=0.17]{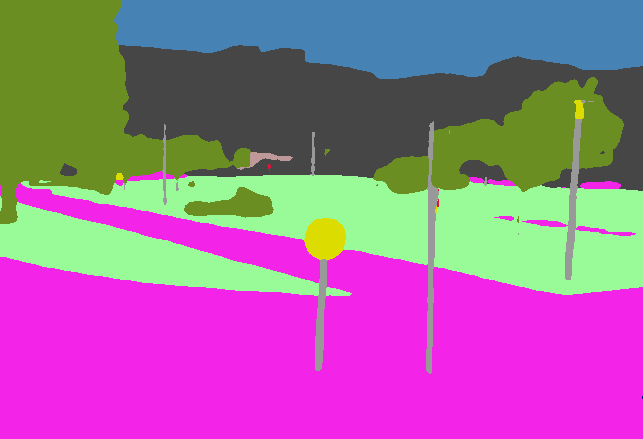}
    \includegraphics[scale=0.17]{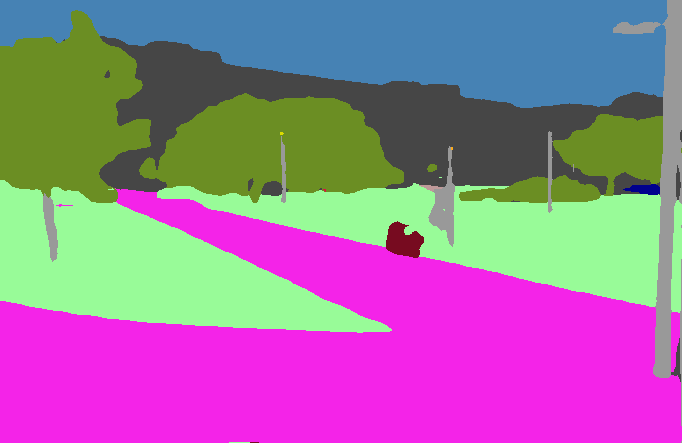}
    \caption{The image on top-left was taken in 2008, the image on top-right was taken in 2019 (source: Google Street View). Observe illumination differences, viewpoint variations and changing objects. Bottom row shows their semantic segmentation results.  Semantic similarity can help to verify/deny the localization result.}
    \label{fig:ebro-campus}
\end{figure}
Long-term visual localization remains a challenging research area since the images taken from the environment can drastically change over time. Any city-scale visual localization system has to handle appearance changes due to weather conditions, seasonal and illumination variations, as well as structural changes such as building facades. Numerous studies have addressed the aspects of long-term localization in the past
\cite{torii201524, toft2020long, seymour2019semantically, naiming2018, sattler2018benchmarking, toft2018semantic, torii2017cvpr, Orhan_2021_ICCV}.
Since semantic information is more robust to such changes (Fig. \ref{fig:ebro-campus}), in our study we utilize semantic segmentation as a side modality at the pose verification step. In other words, we check semantic similarity to verify the poses (retrievals) obtained with the approach that use only RGB image features.
   
To measure semantic similarity, we mainly propose an approach based on self-supervised contrastive learning. 
We train CNN models \cite{khosla2020supervised,chen2021exploring} using a dataset of unlabeled semantic masks. In our case, unlabeled refers to not knowing if two semantically segmented images belong to the same scene or not. We also have a limited size of labeled dataset (query and database images for the same scene) which is used for fine-tuning and testing purposes.
We retrieve top $S$ candidates from the panoramic image database with RGB-only approach and update their similarity scores with semantic features. 
We showed that, this score update with semantic information improves the performance of a state-of-the-art CNN-based visual localization method (SFRS, \cite{ge2020self}).

We also conducted experiments by measuring the semantic similarity at pixel-level, referred to as pixel-wise similarity in the rest of the paper. Since it is the naive approach and easy to implement, we consider it as a baseline. We observed that the pixel-wise similarity can also improve the results of RGB-only approach but when compared to the self-supervised learning with large dataset its gain is marginal.

We summarize our main contributions as follows:
\begin{itemize}
    \vspace{-2mm}
    \item We adopted self-supervised contrastive learning to represent semantic masks. The trained model is used to estimate similarity scores between the semantic contents of different images.
    \vspace{-2mm}
    \item Previous visual localization works utilized semantic information for feature point elimination or for performing localization directly by semantic content. In this work, we take a state-of-the-art image-based localization method and improve it using the proposed semantic similarity estimation approach.
\end{itemize}

The paper is organized as follows. In Section 2, we review the related works. In Section 3, we explain the dataset preparation and demonstrate how we compute and use semantic similarity for pose verification. We present experimental results in Section 4 and conclusions in Section 5.

\section{Related Work}

\textbf{Localization with RGB images.} Before the era of CNN, visual localization was mostly performed by representing images with Bag of Visual Words \cite{philbin2007object}, using SIFT-like hand-crafted descriptors \cite{lowe2004distinctive,bay2006surf}. VLAD (Vector of Locally Aggregated Descriptors) \cite{jegou2011aggregating} does the same task with compact representations that enabled us to use large datasets.

In recent years, CNN-based methods showed great performance on visual localization and image retrieval tasks. One of the first CNN-based approaches was proposed by Razavian \textit{et al.}\cite{razavian2016visual}. They applied the max-pooling function on the last convolution layer of CNN and produced a competitive image representation. Tolias \textit{et al.} \cite{tolias2016particular} improved the previous idea and applied max-pooling to different locations of the convolution layers under different scales. Yi \textit{et al.} \cite{yi2016lift} proposed the LIFT (Learned Invariant Feature Transform) CNN model that consists of detector, orientation estimator, and descriptor parts. Arandjelovic \textit{et al.}  \cite{arandjelovic2016netvlad} proposed NetVLAD that works on geo-tagged images. The proposed model consists of several convolutions and learnable VLAD layers. SFRS \cite{ge2020self}, adopted the backbone of NetVLAD and proposed a training regimen to handle the cases with limited overlap. Instead of using database images as a whole during the training, images are divided into parts and similarity scores are calculated on these parts. By doing so, effect of weakness in GPS labels (position errors) was also alleviated and SFRS outperformed previous works on visual localization.

While the majority of localization methods were applied on standard FOV cameras, there are a few previous works on localization with panoramic images \cite{iscen2017,schroth2011mobile,zamir2010accurate,huang2016fast,Orhan_2021_ICCV}, but these did not exploit semantic information.

\textbf{Semantic-based outdoor localization.} Semantic information is more robust to changes over time and the idea of exploiting semantic content for outdoor visual localization task is not new. We can broadly categorize semantic visual localization methods into 3D structure-based and 2D image retrieval based methods. 3D methods mostly rely on building a 3D model of a scene with structure-from-motion. Stenborg \textit{et al.} \cite{stenborg2018long} performed localization based on the query image’s semantic content when the environment is 3D reconstructed and semantically labeled. In another example, 2D-3D point matches are checked if their semantic labels are also matching \cite{toft2018semantic}.  In our work, we took the 2D approach which retrieves the most similar image to the query. It arguably performs as well as 3D based methods \cite{torii2017cvpr} and less expensive.

Among 2D approaches, Singh and Košecká \cite{singh2012acquiring} utilized semantic layout and trained classifier using semantic descriptor to detect intersection points of the streets. Yu \textit{et al.} \cite{yu2018vlase} proposed a method that utilizes semantic edge features (extracted with CASENet \cite{yu2017casenet}). These edges (e.g. sky-building, building-tree) were converted to vector representation and used for localization. 
Cinaroglu and Bastanlar \cite{cinaroglu2020,cinaroglu2022} trained a CNN model with triplet loss on semantic masks and showed that visual localization can solely be done with semantic features. 
Seymour \textit{et al.} \cite{seymour2019semantically} proposed an attention-based CNN model for 2D visual localization by incorporating appearance and semantic information of the scene. Proposed attention module guide the model to focus on more stable regions. Mousavian \textit{et al.} \cite{mousavian2015semantically} used semantic information to detect man-made landmark structures (e.g. buildings). Feature points not belonging
to man-made objects are considered as unreliable and eliminated. In a similar fashion, Naseer \textit{et al.} \cite{naseer2017semantics} applied a weighting scheme on semantic labels (e.g. increasing
weights for buildings since they are more stable in long term).

These previous works either performed localization only with semantic labels or they used the semantic features to locate where to focus and eliminate unstable regions.
Differently, we check the semantic content in images to validate the retrievals of a state-of-the-art image-based localization method.

\textbf{Contrastive learning.} 
Although its origins date as back as 1990s, contrastive learning has recently gained popularity due to its achievements in self-supervised learning, especially in computer vision \cite{le2020contrastive}. Supervised learning usually requires a decent amount of labeled data, which is not easy to obtain for many applications. With self-supervised learning, we can use inexpensive unlabeled data and achieve a training on a pretext task. Such a training helps us to learn good enough representations. In most cases, a smaller amount of labeled data is used to fine-tune the self-supervised training.

Implemented with Siamese networks, contrastive learning approaches managed to learn powerful representations in a self-supervised fashion. In recently proposed methods, two augmentations of a sample are feed into the networks. The goal of contrastive learning is to learn an embedding space in which similar sample pairs stay close to each other while dissimilar ones are far apart.
While MoCo \cite{he2020momentum} and SimCLR \cite{chen2020simple} use the negative examples directly along with the positive ones, BYOL \cite{grill2020bootstrap} and SimSiam \cite{chen2021exploring} achieved similar performance just with the positive examples (different augmentations of the same sample).
According to the results, not only image classification, but also object detection and semantic segmentation as downstream tasks benefit from self-supervised contrastive learning.

\begin{figure}[htb]
    \centering
    \includegraphics[scale=0.4]{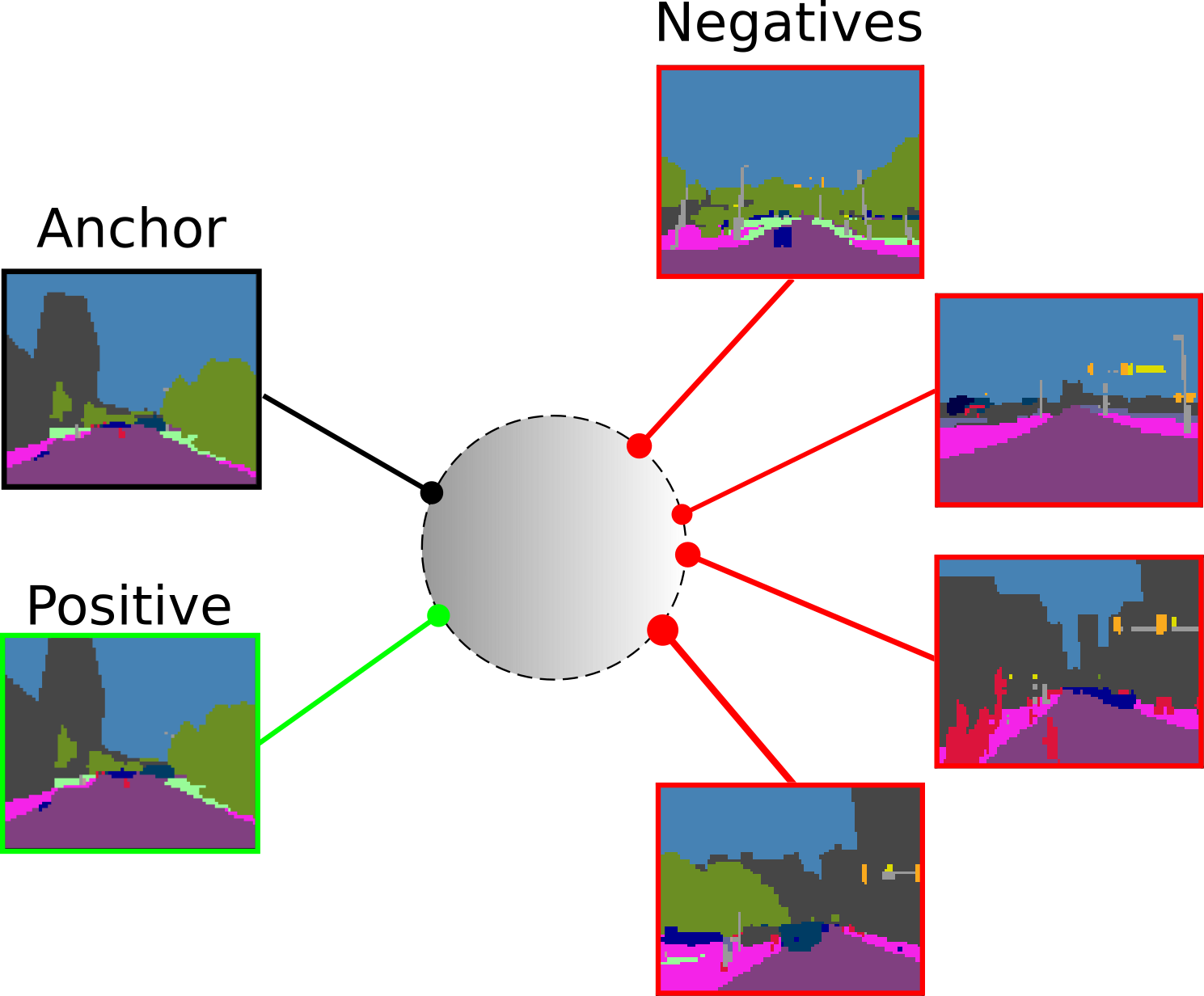}
    \caption{Self-supervised contrastive learning for measuring semantic content similarity. The positive sample is an augmented version of the anchor (we used random crops and small rotations), whereas negative samples belong to different scenes.}
    \label{fig:contrastive_learning}
\end{figure}

In our work, we train a CNN model with a contrastive learning approach to learn similarity scores between semantically segmented images. We have a limited size localization dataset (query and database images with known locations), however it is easy to obtain a large dataset of semantic segmentation masks with unknown locations. Thus, we exploited the power of self-supervised learning to learn from a large unlabeled (no location info) dataset. We mainly used SimCLR \cite{chen2020simple} approach of using both positive and negative samples. To learn similarities in semantic masks,
augmented versions of the anchor are taken as positive,  samples belong to different scenes are taken as negative (Fig. \ref{fig:contrastive_learning}).

\section{Methodology and Dataset}
\subsection{Dataset}
Our dataset consists of images captured in Pittsburgh, PA.
Panoramic images in our dataset were obtained from Google Street View (images of 2019) and downloaded with Street View Download 360 application\footnote{ iStreetView.com}. Query images were taken from UCF dataset \cite{zamir2014image} at locations corresponding to the panoramic images. These queries were also collected via Google Street View but before 2014. This time gap results in seasonal and structural changes (e.g. change of a facade of a building) in addition to illumination variances. Also, a wide camera baseline between the database and query images conforms better to the long-term localization scenario \cite{naiming2018,sattler2018benchmarking,toft2018semantic}.

\begin{figure*}[t!]
    \centering
    \includegraphics[scale=0.069]{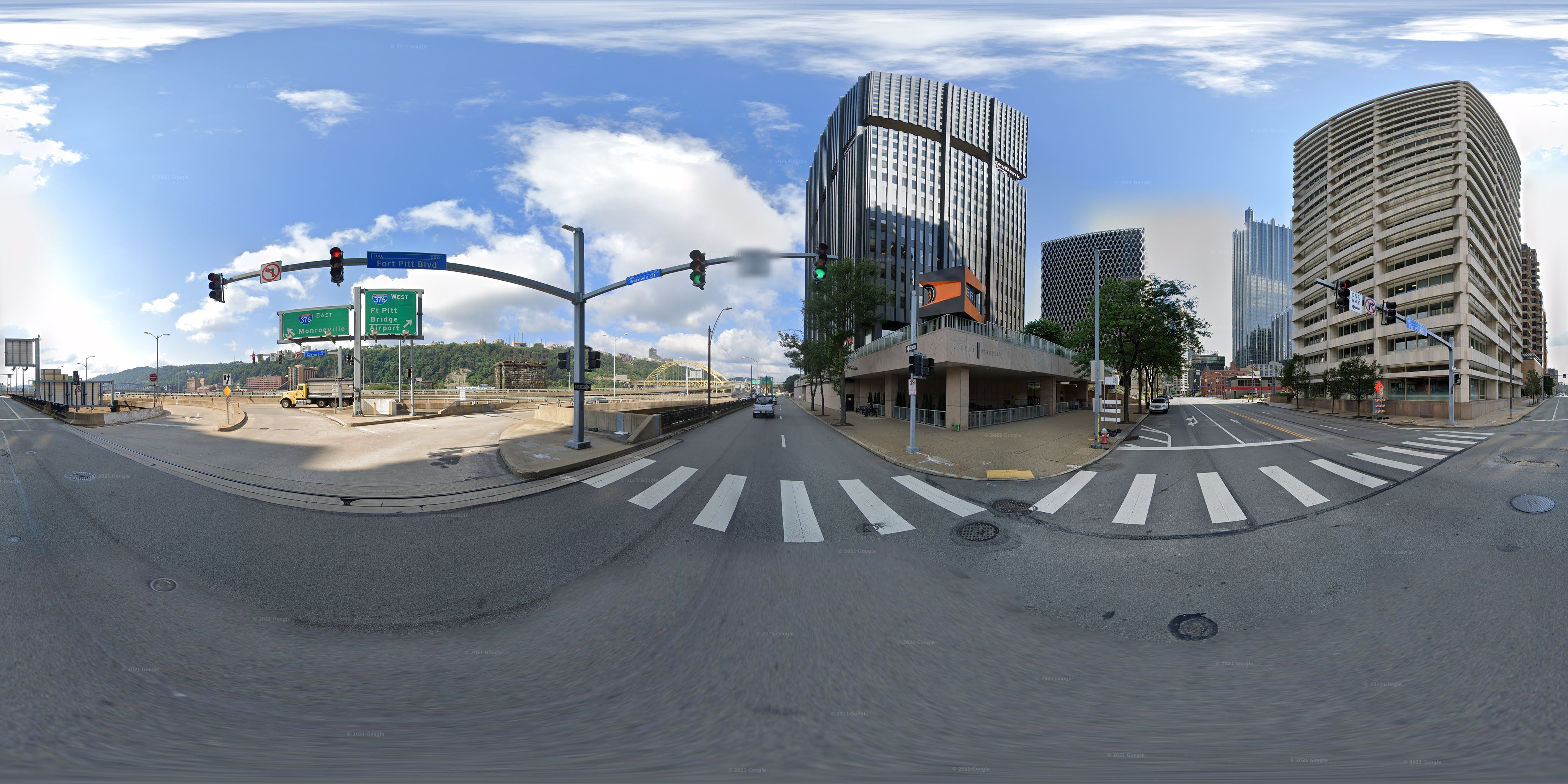}
     \includegraphics[scale=0.43]{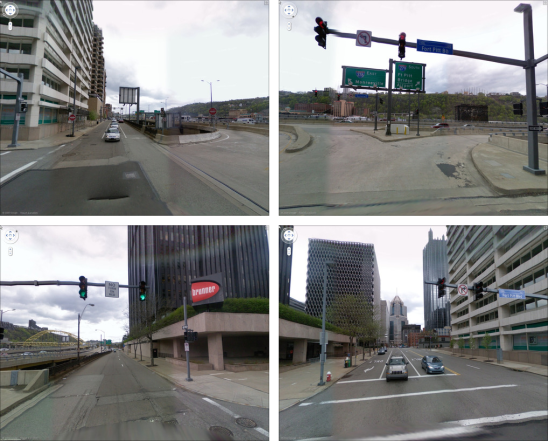}
     \vspace{1mm}\\
     \includegraphics[scale=0.15]{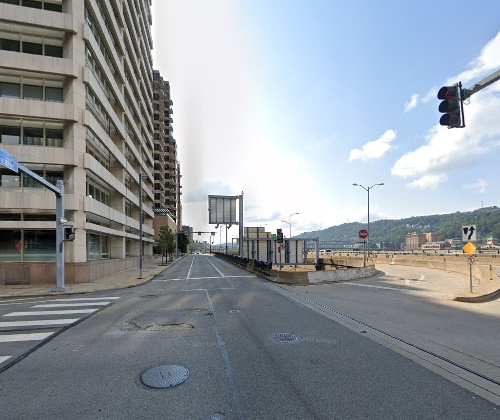}
      \includegraphics[scale=0.15]{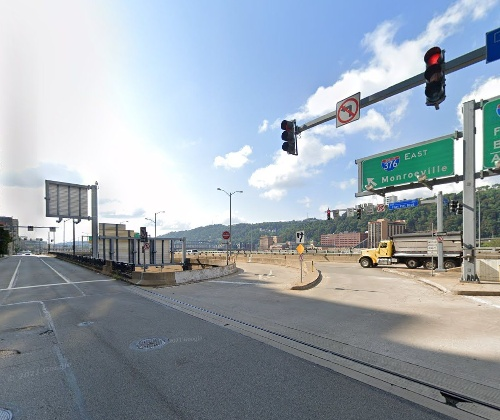}
      \includegraphics[scale=0.15]{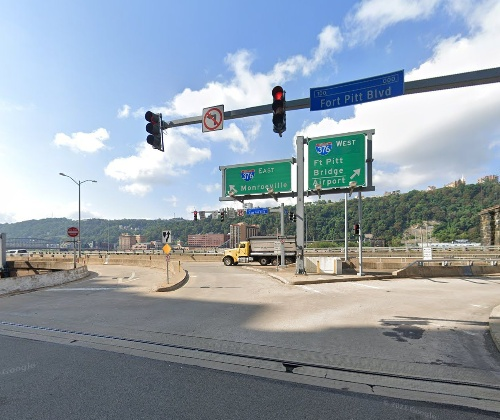}
      \includegraphics[scale=0.15]{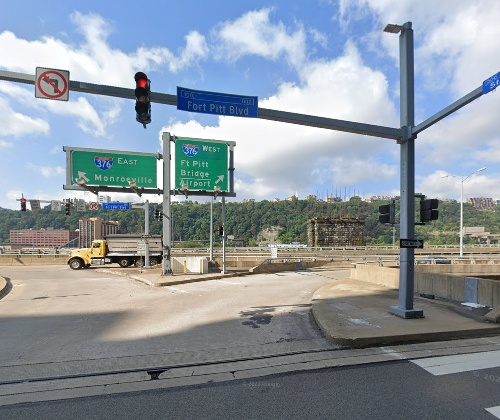}
      \includegraphics[scale=0.15]{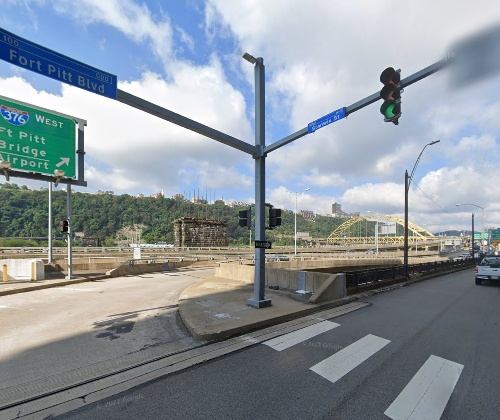}
      \includegraphics[scale=0.15]{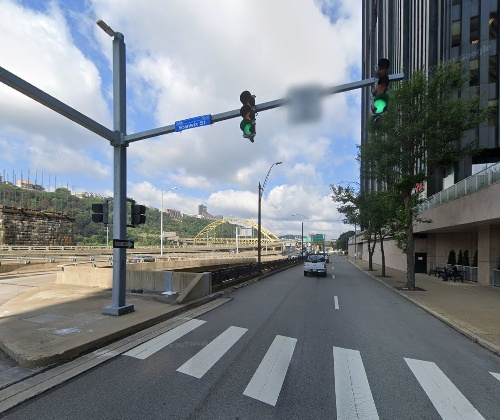}
      \includegraphics[scale=0.15]{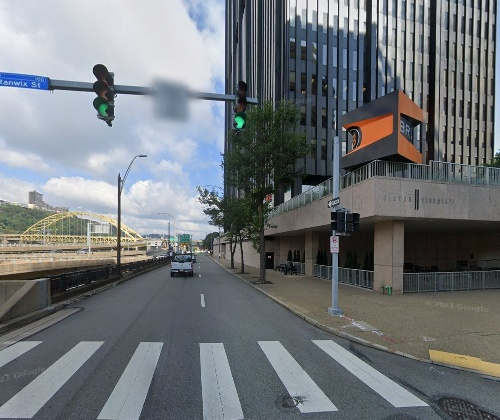}
      \includegraphics[scale=0.15]{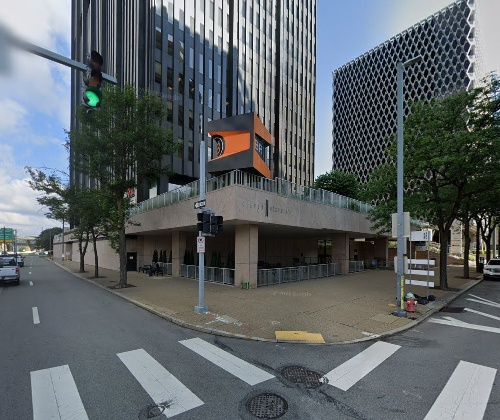}
      \includegraphics[scale=0.15]{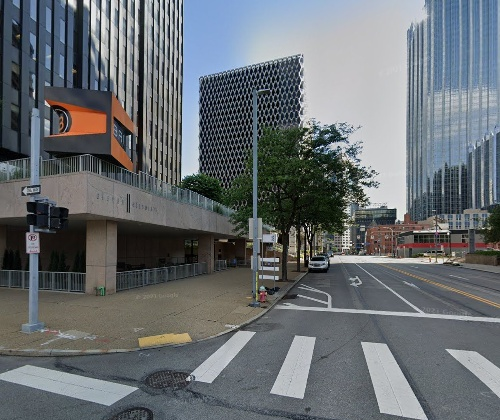}
      \includegraphics[scale=0.15]{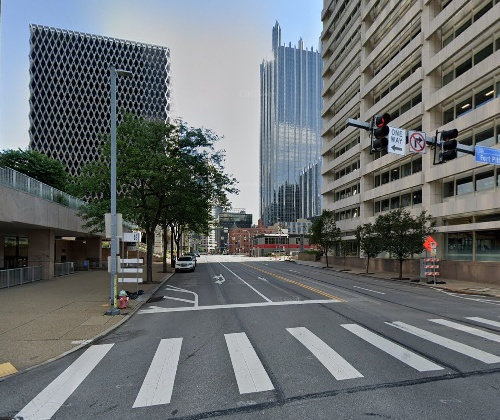}
      \includegraphics[scale=0.15]{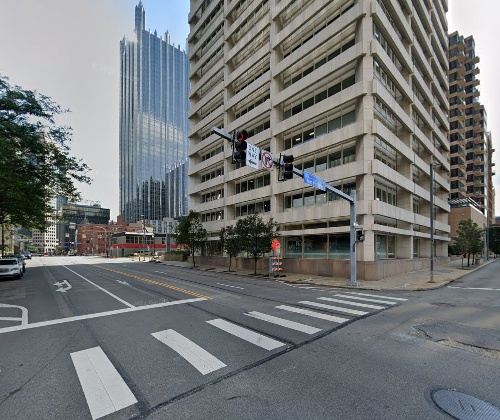}
      \includegraphics[scale=0.15]{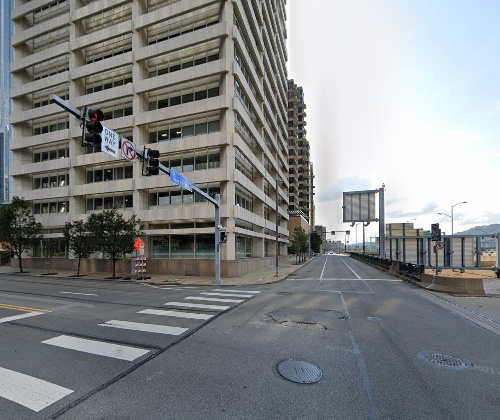}
    \caption{An example set of query and database images taken from the same location. For each location, dataset has one panoramic image  (top-left) and four \ang{90} FOV perspective query images (top-right), captured at different times. To localize, we compare each query image with 12 gnomonic images (bottom two rows) generated from the equirectangular panoramic image.}
    \label{fig:gnomonic_and_query}
\end{figure*}

Query and database images were collected from 123 and 222 different locations respectively. Every query image has a correspondence in the database but not vice versa, i.e. database covers a larger area. Query set consists of 123x4=492 non-overlapping perspective images (\ang{90} FOV each). Database consists of 222x12=2674 images (each panoramic image is represented with 12 gnomonic images).
Each gnomonic image also has \ang{90} FOV and it overlaps \ang{60} with the next one. 
Please see Fig. \ref{fig:gnomonic_and_query} for examples.

\subsection{Searching perspective query images in a panoramic image database}

We search query images in the 12-gnomonic image database that is generated from equirectangular panoramic images. Images collected from the same location at different times not only contain appearance changes but also may suffer from limited view overlap due to position and orientation shifts. 
Even though we use 12 consecutive gnomonic projections on the database side to increase the chance of a good overlap between query and database (there is a \ang{15} viewpoint difference in worst case and \ang{7.5} on average), we can not guarantee a perfect overlap. Several successful methods were proposed to cope with both appearance and viewpoint changes. \cite{ge2020self,arandjelovic2016netvlad, tolias2016particular}. Most recently, SFRS \cite{ge2020self} outperformed other methods on visual localization benchmark datasets taking advantage of image-to-region similarities. Thus, without loss of generality, we take SFRS \cite{ge2020self} as the state-of-the-art image-based method and we verify its retrieval results using semantic similarity. 

\subsection{Computing Semantic Similarity}

We first automatically generate a semantic mask for each image in our dataset using a well-performing CNN model \cite{sun2019high}. The model we employed was trained on Cityscapes \cite{cordts2016cityscapes}, which is an urban scene understanding dataset consists of $30$ visual classes, such as building, sky, road, car, etc.

Given a semantic mask, obtaining the most similar result among the alternatives is not a trivial task. SIFT-like features do not exist to match. Moreover, two masks of the same scene is far from being identical not only because of changing content but also due to camera position and viewpoint variations. We have tried geometric methods \cite{rocco2017} to fit one image into the other one prior to computing semantic similarity, however they did not succeed. 
Thus, we propose a trainable semantic feature extractor for pose verification which is trained using correct and incorrect pose matchings.

Before presenting the trainable approach, we explain pixel-wise similarity approach which measures the semantic similarity at pixel-level. Since it is the naive approach and easy to implement, we see this as a baseline method of semantic pose verification.

\vspace{2mm}
\textbf{Pixel-wise Similarity.} 
In this first approach, we calculate pixel-by-pixel similarity between query and database semantic masks:
\begin{equation}
    \textit{pixel-wise similarity}=\frac{\displaystyle \sum_{i=1}^{m} \sum_{j=1}^{n}  sim(Q_{(i,j)},D_{(i,j)})} {\displaystyle m \cdot n}
    \label{eq:pixel-acc}
\end{equation}

\noindent where $sim(a,b)$ is equal to 1 if $a=b$, 0 otherwise. $Q$ represents the query image's mask and $D$ represents the database image's mask, both having size $m\times n$, $(i,j) \in \{1,...,m\} \times \{1,...,n\}$. A pixel is considered as a matching pixel if $Q_{(i,j)} = D_{(i,j)}$ and it increases similarity. 

\vspace{2mm}
\textbf{Trainable Feature Extractor.} 
We use self-supervised contrastive learning approach in our work since large amount of semantic masks can easily be obtained for a self-supervised training. In our setting, semantic masks are obtained with a well-performing segmentation model \cite{sun2019high} from 3484 images randomly taken from UCF dataset \cite{zamir2014image}. We do not need groundtruth masks, since a successful estimation is enough to compute semantic similarity.

\begin{figure}[t]
    \centering
    \includegraphics[scale=0.45]{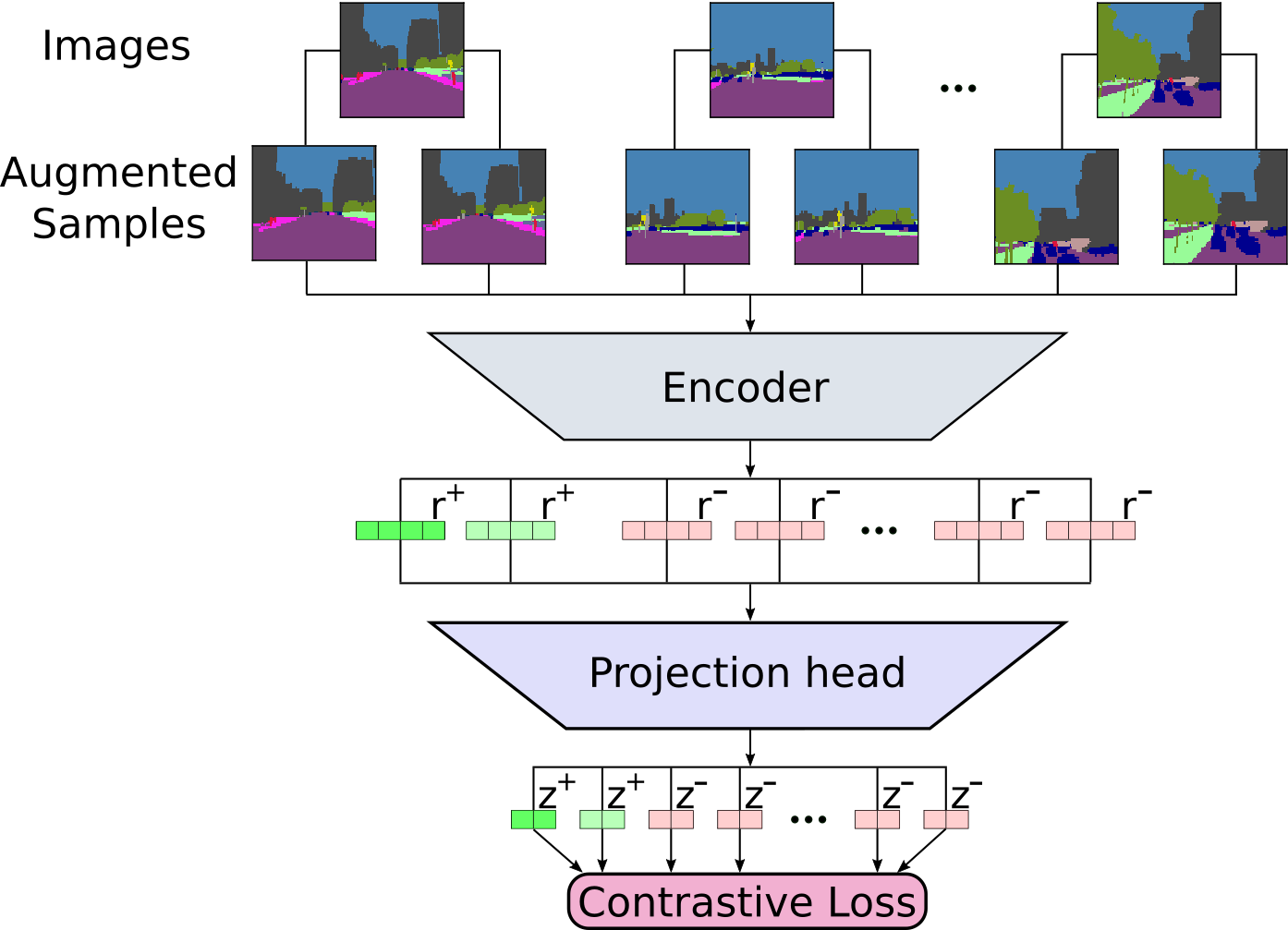}
    \caption{Illustration of training a CNN model with self-supervised contrastive loss on a dataset that consists of semantically segmented masks.}
    \label{fig:contrastive_learning_arc}
\end{figure}

We used SimCLR \cite{chen2020simple}
as our contrastive learning model and trained a ResNet-18 as the encoder. In our setting, encoder network (Fig. \ref{fig:contrastive_learning_arc}) produces $r= Enc(x) \in R^{512}$ dimensional features, projection network produces $z= P roj(r) \in R^{2048}$ dimensional features. We resized semantic mask to $64 \times 80$ resolution (due to GPU memory limitation) and used two different data augmentation methods during the training: random resized crop and random rotation. We set lower bound of random crop parameter as $0.6$, which means that cropped mask covers at least 60\% area of the original mask. We set maximum rotation parameter as \ang{3}, since severe rotations are not expected between query and database images. Augmentation of semantic masks is visualized in Fig. \ref{fig:contrastive_learning_arc}. Following \cite{khosla2020supervised,chen2020simple} we use the contrastive loss given in Eq. \ref{eq:contrastive_loss}. This is a categorical cross-entropy loss to identify the positive sample amongst a set of negative samples (inspired from InfoNCE \cite{oord2018}).

\begin{equation}
    L^{self}= \sum_{i \in I}L_{i}^{self} = -\sum_{i \in I}\mathrm{log}\frac{\mathrm{exp}(z_i \cdot z_{j(i)}/ \tau)}{\displaystyle\sum_{a \in A(i)} \mathrm{exp}(z_i  \cdot z_{a(i)}/\tau)}
    \label{eq:contrastive_loss}
\end{equation}

$N$ images are randomly taken from the dataset. Thus, the training batch consists of 2$N$ images to which data augmentations are randomly applied. 
Let $i \in I \equiv \{1...2N\}$ be the index of an arbitrary augmented sample, then $j(i)$ is the index of the other augmentation of the same original image.
$\tau \in R^{+}$ is a scalar temperature parameter, $\cdot$ represents the dot product, and $A(i) \equiv I - \{i\}$. We call index $i$ the anchor, index  $j(i)$ is the positive, and the other $2(N-1)$ indices as negatives. 
%There is 1 positive pair and 2N-2 negative pairs. 
The denominator has a total of $2N-1$ terms (one positive and $2N-2$ negatives).

CNN model, trained as explained above, is now ready to produce a similarity score when two semantic masks (one query and one database) are given. Similarity score is used to update the scores of RGB-only method (Section \ref{updatesimilarity}).

After self-supervised training, same network can be fine-tuned with a labeled dataset (query and database segmentation masks for the same scene). For this purpose, we prepared a dataset of 227 query images with their corresponding database panoramic images. Not surprisingly, it is much smaller than the self-supervised training dataset. Here, common practice in literature is that the projection head (Fig.\ref{fig:contrastive_learning_arc}) is removed after pretraining and a classifier head is added and trained with the labeled data for the downstream task. However, since our pretraining and downstream tasks are the same (estimating similarity of two input semantic masks), we do not place a classifier head, but we retrain the network (partially or full).

\subsection{Updating Retrieval Results with Semantic Similarity}
\label{updatesimilarity}

\begin{figure*}[th!]
    \centering
    \includegraphics[scale=0.80]{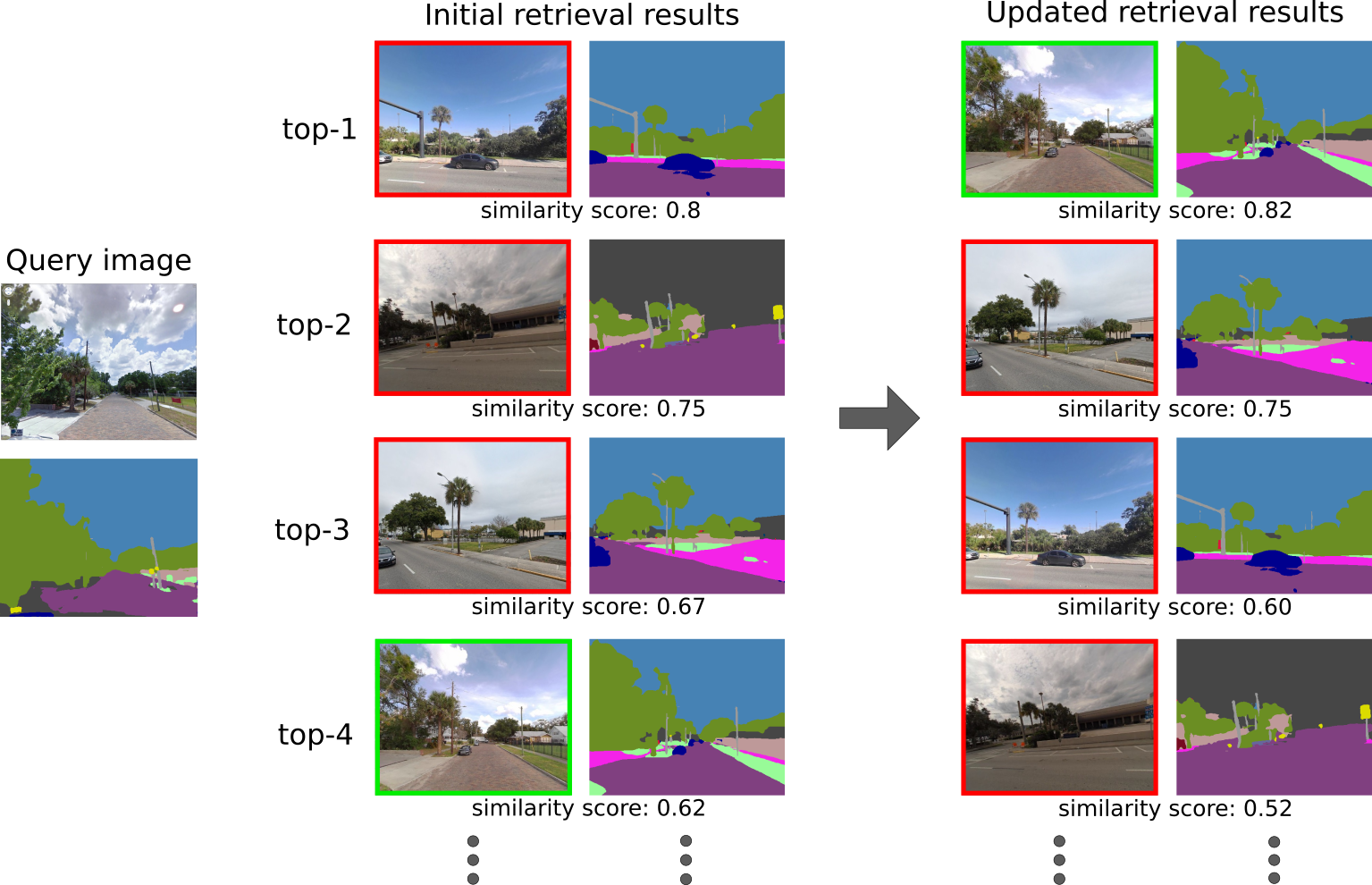}
    \caption{An example set of retrieval results from the database where RGB-only method fails to correctly localize but the method merging RGB and semantic scores correctly localizes. Green rectangle indicates correct match, which increased to top rank after score update.}
    \label{fig:semantic_similarity}
\end{figure*}

We first normalize RGB-only \cite{ge2020self} and semantic similarity (pixel-wise similarity or trainable feature extractor) scores between $[-1,+1]$ and then merge them with a weight coefficient ($W$) to obtain the updated similarity score: 
\begin{equation}
    \textit{updated-score}_i = \textit{rgb-score}_i+\textit{W}\cdot \textit{semantic-score}_i 
    \label{eq:rgb_scores_update}
\end{equation}
where $i$ is the index within the top $S$ candidates for each query. We do not update the scores of every database image, but only interested in the top $S$ database candidates for each query, since these are already obtained by a state-of-the-art image-based localization method.  
%$i=1,..., S$. 
We set $S$=10 in our experiments. A real (successful) example of similarity score update is shown in Fig.\ref{fig:semantic_similarity}.

While updating the similarity scores of $S$ candidates, we also update the similarity scores of gnomonic images coming from the same panorama (neighbours of the retrieved gnomonic views also have potential to benefit from semantic similarity).

%We set $0.10$ for pixel-wise similarity, $0.25$ for trainable feature extractor methods. 
To decide on $W$ value, we employed a validation set of query-database image pairs which are not included in the test set. We selected $W$ values with the highest localization performance, separately for pixel-wise similarity and trainable semantic feature extractor approaches. Effect of altering $W$ was examined with experiments in Section \ref{ablation}.

\section{Experimental Results}

%\subsection{Training details}
Our self-supervised contrastive learning uses 3484 images randomly taken from UCF dataset \cite{zamir2014image} and not coincide with the test set. We used stochastic gradient descent optimizer with initial learning rate = $0.05$. Temperature parameter ($\tau$) was taken as $0.07$ and batch size ($2N$) as 174.

%\subsection{Visual-based Localization Results}
With experiments, we compare the localization performances of three approaches. First is the state-of-the-art visual localization with RGB image features \cite{ge2020self}, second approach is updating this RGB-only method's scores with pixel-wise semantic similarity, and third is updating RGB-only method's scores with the similarity given by trainable semantic feature extractor (with SimCLR\cite{chen2020simple} self-supervised training scheme). We also conducted experiments with models fine-tuned with labeled dataset after self-supervised training. Those results will be presented in Section \ref{ablation}.

\begin{figure*}[th!]
    \centering
    \includegraphics[scale=0.80]{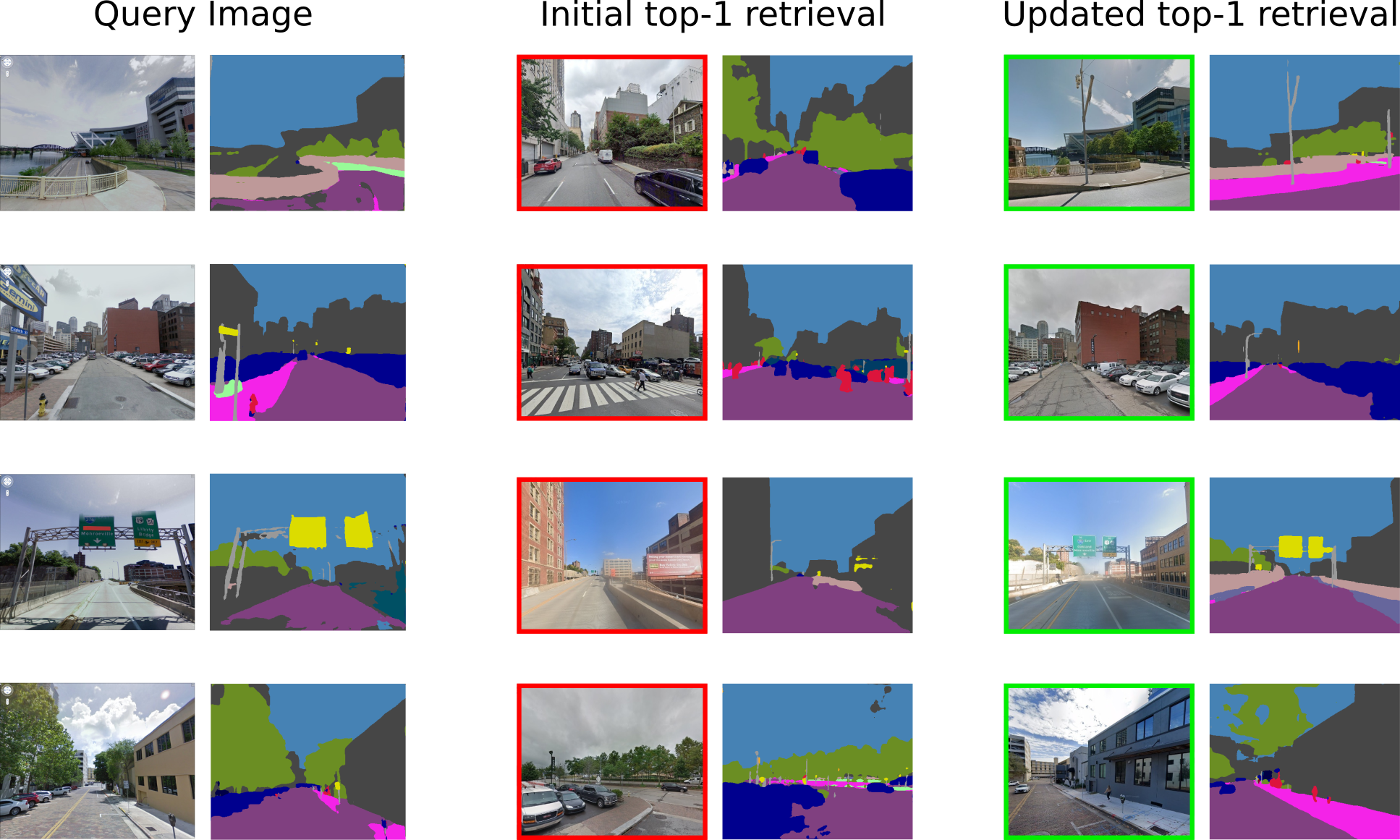}
    \caption{Example retrieval results in which utilizing semantic similarity scores at pose verification improved the localization performance of the RGB-only model. Query images are in the first column, top-1 retrieval results are in the middle column, and updated top-1 retrieval results with trainable semantic feature extractor are presented in the last column. Utilizing semantic similarity moved up the correct candidates in ranking when semantic contents of query and database images are  similar. Distinctive objects (e.g. traffic signs) help to correctly localize query images with semantic information (third row of the figure). In some cases, localization was improved even the semantic masks of the images contain labeling errors (last row of the figure).}
    \label{fig:semantic_cont}
\end{figure*}

\begin{figure}[htb]
    \centering
      \includegraphics[scale=0.31]{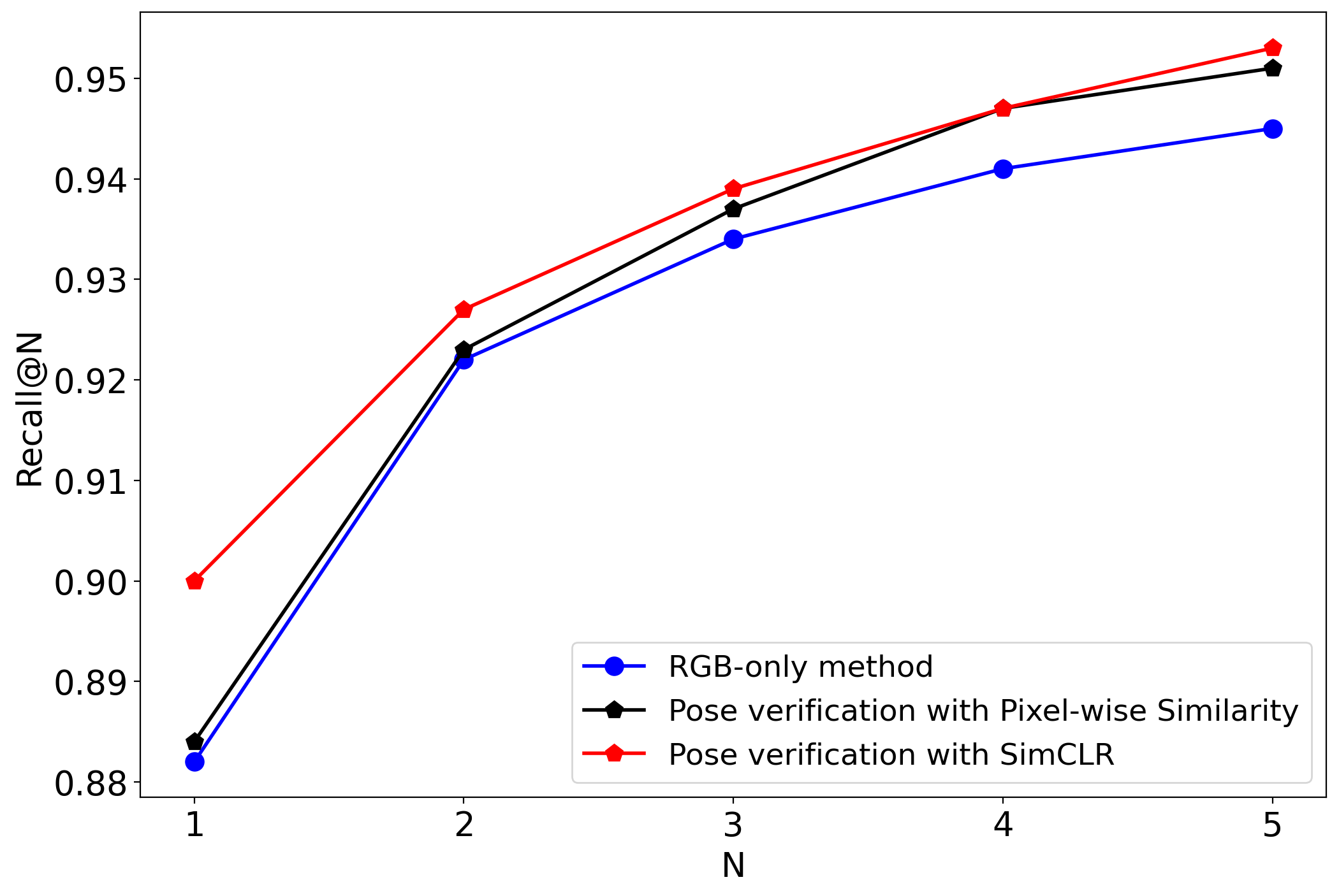}
\caption{Visual-based localization results of RGB-only method, pose verification with pixel-wise similarity, and pose verification with self-supervised learning (SimCLR).}
\label{fig:visual_localization}
\end{figure}

\begin{figure}[htb]
    \centering
      \includegraphics[scale=0.31]{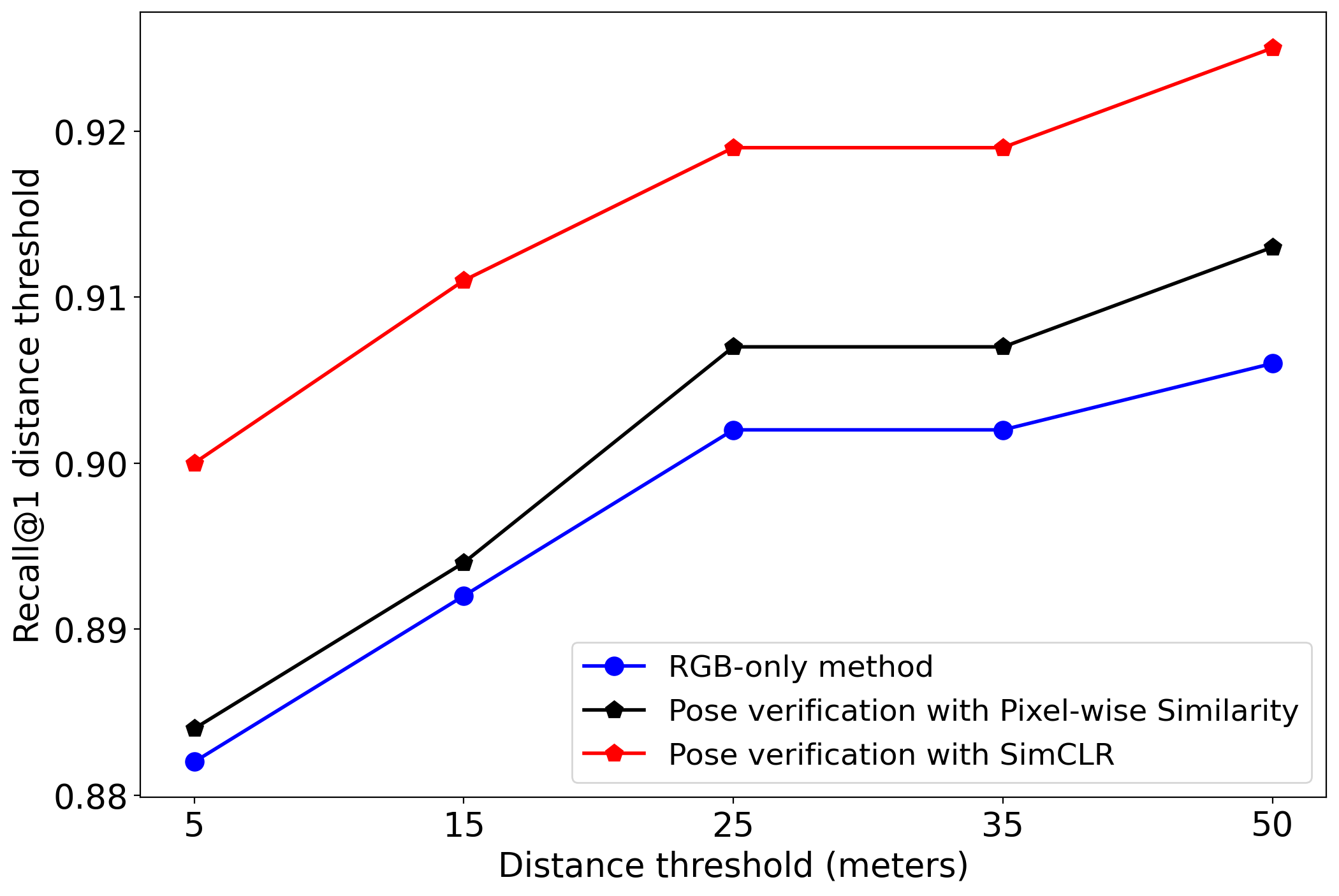}
\caption{Localization results of three approaches where N=1 with different distance thresholds.}
\label{fig:metric_localization}
\end{figure}

Performances of different approaches are compared with Recall@N metric. According to this metric, a query image is considered as correctly localized if the distance between the query and any retrieved database images in top-N is smaller than the metric distance threshold. The threshold was set as 5 meters in our experiments. Since the test dataset is prepared so that a database image is taken at the location of each query image, ideally all the queries can be localized with 5-meter threshold. 
Results in Fig.\ref{fig:visual_localization} show that the semantic pose verification is useful for all cases and it improves Recall@1 of RGB-only model by \%2 when the proposed trainable semantic feature extractor is trained in a self-supervised fashion with SimCLR. 
In addition to Recall@N plots in Fig.\ref{fig:visual_localization},
%metric at 5 meter distance threshold, 
we provide Recall@1 performances with varying distance thresholds in Fig.\ref{fig:metric_localization}. We observe that pose verification with trainable feature extractor continues to outperform RGB-only approach and pose verification with pixel-wise similarity approach for increasing distance thresholds. Lastly, Fig.\ref{fig:semantic_cont} shows several examples where the proposed semantic pose verification improved the results. 

\subsection{Ablation Study}
\label{ablation}
Experimental results given so far were obtained with self-supervised contrastive learning without any fine-tuning, minimum crop ratio was used as $0.6$ and $W$ was set to $0.25$ as suggested values. Now, we present our fine-tuning results and additional experiments to investigate how much our approach is sensitive to crop ratio and $W$ parameters. 

Table \ref{tab:ablation_study} compares self-supervised learning results with alternatives where the model is fine-tuned with a
labeled dataset which corresponds to 227 query semantic masks and their actual corresponding semantic masks in the database. Three different fine-tuning schemes were tested: last two dense layers were retrained, two new dense layers were added and trained, all layers of network were trained. None of them improved self-supervised training and only fine-tuning all layers kept a similar performance. It should be noted that our labeled dataset is much smaller than the unlabeled dataset (227 $\ll$ 3484). Another reason for not improving with fine-tuning could be the fact that our main and downstream tasks are the same, i.e. scoring similarity between two semantic masks. Whereas, successful examples of fine-tuning in literature contains placing a classifier head and training for a different downstream task like image classification or object detection.

\begin{table*}[htb]
\centering
\caption{Self-supervised model is compared with fine-tuned models. Results were obtained with semantic score weight $W=0.25$.}
{\small
\begin{tabular}{c|c|cccc}
\hline
\multirow{2}{*}{\textbf{Training Methods}} & \multicolumn{5}{c}{\textbf{Recall@N}}\\% \cline{2-6}
& \multicolumn{1}{l}{\textbf{N=1}} & \multicolumn{1}{l}{\textbf{N=2}} & \multicolumn{1}{l}{\textbf{N=3}} & \multicolumn{1}{l}{\textbf{N=4}} & \multicolumn{1}{l}{\textbf{N=5}} \\
\hline

\multicolumn{1}{l|}{Only self-supervised training} &  \multicolumn{1}{c}{\textbf{0.900}} & \multicolumn{1}{c}{0.927} & \multicolumn{1}{c}{0.939} & \multicolumn{1}{c}{0.947} & \multicolumn{1}{c}{0.953} \\
% \hline

\multicolumn{1}{l|}{Fine-tuning last two dense layers} &  \multicolumn{1}{c}{0.890} & \multicolumn{1}{c}{0.927} & \multicolumn{1}{c}{0.937} & \multicolumn{1}{c}{0.943} & \multicolumn{1}{c}{0.945} \\
% \hline

\multicolumn{1}{l|}{Adding two new dense layers} &  \multicolumn{1}{c}{0.892} & \multicolumn{1}{c}{0.927} & \multicolumn{1}{c}{0.935} & \multicolumn{1}{c}{0.947} & \multicolumn{1}{c}{0.947} \\
% \hline

\multicolumn{1}{l|}{Fine-tuning all layers} & \multicolumn{1}{c}{\textbf{0.900}} & \multicolumn{1}{c}{0.925} & \multicolumn{1}{c}{0.933} & \multicolumn{1}{c}{0.943} & \multicolumn{1}{c}{0.945} \\
\hline

\end{tabular}
}
\label{tab:ablation_study}
\end{table*}

\begin{table}[htb]
    \centering
    \caption{
    Effect of the minimum crop ratio parameter in data augmentation on localization performance.}
    {\small
    \begin{tabular}{c|ccccc} % |c|c|c|c|c|c|
    \hline
    \multirow{2}{*}{\textbf{\shortstack[c]{Crop\\ Ratio}}} & \multicolumn{5}{c}{\textbf{Recall@N}}\\ %\cline{2-6}
    & \multicolumn{1}{l}{\textbf{N=1}} & \multicolumn{1}{l}{\textbf{N=2}} & \multicolumn{1}{l}{\textbf{N=3}} & \multicolumn{1}{l}{\textbf{N=4}} & \multicolumn{1}{l}{\textbf{N=5}} \\
    \hline
    0.90 & 0.888 & 0.921 & 0.931 & 0.939 & 0.939 \\
    0.80 & 0.892 & 0.927 & 0.937 & 0.943 & 0.947 \\
    0.70 & 0.896 & 0.931 & 0.941 & 0.943 & 0.947 \\
    0.60 & \textbf{0.900} & 0.927 & 0.939 & 0.947 & 0.953 \\
    0.50 & 0.898 & 0.929 & 0.937 & 0.941 & 0.943 \\
    0.40 & 0.894 & 0.931 & 0.937 & 0.943 & 0.947 \\
    0.30 & 0.894 & 0.927 & 0.933 & 0.943 & 0.947 \\
    \hline
    \end{tabular}
    }
    \label{tab:crop_ration}
\end{table}

\begin{table}[htb]
    \centering
    \caption{Effect of the weight of semantic similarity score ($W$) on localization performance.}
    {\small
    \begin{tabular}{c|ccccc} % |c|c|c|c|c|c|
    \hline
    \multirow{2}{*}{\textbf{\shortstack[c]{Semantic\\ Weight}}} & \multicolumn{5}{c}{\textbf{Recall@N}}\\% \cline{2-6}
    & \multicolumn{1}{l}{\textbf{N=1}} & \multicolumn{1}{l}{\textbf{N=2}} & \multicolumn{1}{l}{\textbf{N=3}} & \multicolumn{1}{l}{\textbf{N=4}} & \multicolumn{1}{l}{\textbf{N=5}} \\
    \hline
    0.10 & 0.884 & 0.925 & 0.935 & 0.945 & 0.951 \\
    0.15 & 0.884 & 0.929 & 0.937 & 0.943 & 0.951 \\
    0.20 & 0.892 & 0.929 & 0.937 & 0.949 & 0.953 \\
    0.25 & \textbf{0.900} & 0.927 & 0.939 & 0.947 & 0.953 \\
    0.30 & \textbf{0.902} & 0.929 & 0.937 & 0.945 & 0.953 \\
    0.35 & \textbf{0.902} & 0.931 & 0.937 & 0.943 & 0.949 \\
    0.40 & \textbf{0.900} & 0.929 & 0.937 & 0.945 & 0.949 \\
    0.45 & 0.888 & 0.925 & 0.939 & 0.945 & 0.947 \\
    0.50 & 0.882 & 0.923 & 0.939 & 0.945 & 0.947 \\
    \hline
    \end{tabular}
    }
    \label{tab:semant_weight_coefficient}
\end{table}

We have also evaluated another self-supervised training approach, SimSiam \cite{chen2021exploring}, however its performance was worse than SimCLR. Thus, we excluded it from our ablation study.

Table \ref{tab:crop_ration} presents the effect of minimum crop ratio parameter used in data augmentation module. Values fluctuate in a close range for N=\{2,..,5\}, however, Recall@1 is highest for $0.6$ and performance gradually drops as we increase or decrease the minimum crop ratio. This is in accordance with the finding in \cite{tian2020} that there is a reverse-U shaped relationship between the performance and the mutual information within augmented views. When crops are close to each other (high mutual information, e.g. crop ratio=$0.9$) the model does not benefit from them much. On the other hand, for low crop ratios (low mutual information) model can not learn well since views look quite different from each other. Peak performance stays somewhere in between.

Lastly, Table \ref{tab:semant_weight_coefficient}
shows the results of the experiments with different semantic weight coefficient ($W$). We understand that success is not specific to $W=0.25$ and pose verification works equally well for values between $0.25$ and $0.40$.

\section{Conclusion}
In this work, we localize perspective query images in a geo-tagged database of panoramic images. We take advantage of semantic segmentation masks due to their robustness to long-term changes. Semantic similarity is measured via pixel-wise similarity and trainable feature extractors. Experimental results showed that utilizing semantic similarity at pose verification step contributed to visual localization performance of a state-of-the-art method \cite{ge2020self}.
Gained improvement is due to the more stable semantic content and does not depend on which localization method used to obtain initial retrieval results. Thus, other RGB image based visual localization methods can be improved in the same manner.

We also conclude that pose verification with a CNN model, which exploits self-supervised contrastive learning, performs better than using pixel-wise similarity between masks.
This confirms the potential of self-supervised models for representation learning when there is a limited amount of labeled data.

There are works that search the query image within the panoramic image instead of using gnomonic views (e.g. \cite{iscen2017,Orhan_2021_ICCV}). Also, semantic segmentation masks can be obtained for panoramic images of street views \cite{orhan2021panoramicsemantic}. A future work may be extending our effort to compute semantic similarity directly from panoramic semantic masks.

\section*{Acknowledgments}
\vspace{-2mm}
This work was supported by the Scientific and Technological Research Council of Turkey (TÜBİTAK) under Grant No. 120E500 and also under 2214-A International Researcher Fellowship Programme.
%%%%%%%%% REFERENCES
{\small
\bibliographystyle{ieee_fullname}
\bibliography{ref}
}

\end{document}